\documentclass{article}
\usepackage{spconf,amsmath,graphicx}
\usepackage{algorithm}
\usepackage{algorithmic}
\usepackage{rotating}
\usepackage{hyperref}
\usepackage{multirow}
\usepackage{booktabs}
\usepackage{longtable}

\title{Guided learning for weakly-labeled semi-supervised Sound event detection}
\name{Liwei Lin$^{1,2}$, Xiangdong Wang$^{1,\dagger}$, Hong Liu$^1$, Yueliang Qian$^1$}
\address{
  $^1$Bejing Key Laboratory of Mobile Computing and Pervasive Device,\\
  Institute of Computing Technology, 
  Chinese Academy of Sciences, Beijing, China\\
  $^2$University of Chinese Academy of Sciences, Beijing, China\\
  \{linliwei17g, xdwang, hliu, ylqian\}@ict.ac.cn}
\begin{document}
%
\maketitle
\begin{abstract}
We propose a simple but efficient method termed Guided Learning for weakly-labeled semi-supervised sound event detection (SED). There are two sub-targets implied in weakly-labeled SED: audio tagging and boundary detection. Instead of designing a single model by considering a trade-off between the two sub-targets, we design a teacher model aiming at audio tagging to guide a student model aiming at boundary detection to learn using the unlabeled data. The guidance is guaranteed by the audio tagging performance gap of the two models. In the meantime, the student model liberated from the trade-off is able to provide more excellent boundary detection results. We propose a principle to design such two models based on the relation between the temporal compression scale and the two sub-targets. We also propose an end-to-end semi-supervised learning process for these two models to enable their abilities to rise alternately. Experiments on the DCASE2018 Task4 dataset show that our approach achieves competitive performance.

\end{abstract}
\section{Introduction}
\label{sec:intro}

Sound event detection (SED) consists in recognizing the presence of sound events in segments of audio and detecting their onsets as well as offsets. Traditional approaches of SED requires a large amount of training data with detailed annotations \cite{cakir2015polyphonic, parascandolo2016recurrent,cakir2017convolutional}. Due to the high cost of large-scale detailed annotations, weakly-labeled semi-supervised SED, which employs a small number of weak annotations that only indicates the presence of the event classes together with a large amount of unlabeled data, has become a new focus in the research on SED \cite{zhang2012semi,hua2013sted,shi2019semi}. Especially, DCASE2018 Task 4 \cite{Serizel2018} and DCASE2019 Task 4 \cite{turpault:hal-02160855} have launched challenges on weakly-labeled semi-supervised SED in domestic environments.

There are two sub-targets implied in weakly-labeled SED: audio tagging to determine the occurrences of the event classes and boundary detection to determine the timestamps of all the occurring events. Due to the excellent performance of deep neural networks, most recent weakly-labeled SED systems are based on deep neural networks such as CNN \cite{su2017weakly,mcfee2018adaptive,kumar2018knowledge,kong2019sound} and CRNN \cite{jiakai2018mean}. However, as discussed in \cite{long2015fully}, when designing deep neural networks, there is always a trade-off between the two sub-targets: larger compression scale of the high-level feature map of neural networks tends to achieve more accurate classification performance (audio tagging) but leads to less detail information (for boundary detection). Therefore, previous studies on weakly-labeled semi-supervised SED apply general semi-supervised methods such as Mean Teacher \cite{tarvainen2017mean,jiakai2018mean} and Triplet loss \cite{turpault2019semi} to a well-designed model. The model with a moderate temporal compression scale seeks a trade-off between the two sub-targets and these semi-supervised methods focus on optimizing the direct training target (audio tagging) using unlabeled data. However, the trade-off implied in the model architecture still limits the boundary detection capability of the SED system.

In this paper, we propose a teacher-student framework named Guided Learning for weakly-labeled semi-supervised SED. Unlike Mean Teacher \cite{tarvainen2017mean}, where the teacher model is the temporal weighted average of the student model so that the two models are homologous, the proposed method uses two different models pursuing the two sub-targets respectively. The two models are trained synchronously and forced to learn from the unlabeled data with tags generated by each other. In this way, the performance gap on audio tagging between the two models enables the teacher model to guide the student model to learn using unlabeled data. When the student model performs well enough on audio tagging, it can in turn help to fine-tune the teacher model. Therefore, the abilities of these two models rise alternately during the training. This separation of the sub-targets has two advantages: 1) the student model is liberated from the trade-off required for a well-designed model in other semi-supervised learning methods and expected to achieve better boundary detection performance; 2) it provides a direct way of reusing the STOA models of the two sub-targets, which means that new techniques of audio tagging can be used without or with little concern of boundary detection, and vice versa. Experimental results on the DCASE2018 Task 4 dataset show that our approach achieves competitive performance.

\begin{figure}[t]
\centering
\includegraphics[width=\linewidth]{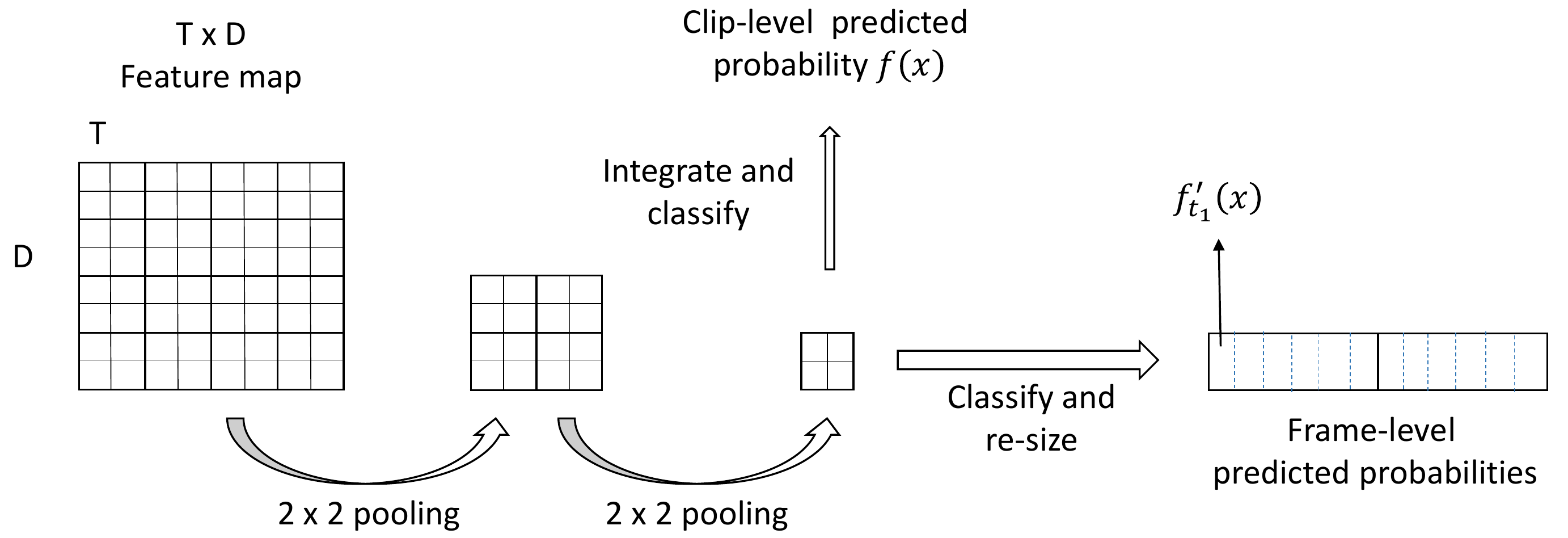}
\vskip -0.1in
\caption{The general design of the neural network. 3 grids on the left represent intermediate feature maps. Several pooling layers lead to the temporal compression. When the classification probabilities are restored to the original size for frame-level prediction, most of the detailed information lost.}
\label{fig3}
\vskip -0.15in
\end{figure}

\section{Guided learning}
\label{sec:GL}

\subsection{The design principle of a more professional teacher model and a more promising student model}
In this section, we describe in detail the trade-off mentioned in Section~\ref{sec:intro}, introduce why and how we design the two models.

Assuming that $x$ is the input feature of the neural network and to simplify the description, we formulate the problem as a binary detection problem of each single event category. Let $f(x)$ denotes the clip-level predicted probability output by the neural network for input $x$ from an audio clip, then the clip-level prediction is:
\begin{equation}
    \mathbf{\phi}(f(x))=
    \left\{\begin{matrix}
1,&f(x)\geq \alpha \\ 
0,&otherwise
\end{matrix}\right.
\label{eq1}
\end{equation}

Similarly, assuming that $f_{t}^{'}(x)$ denotes the $t^{th}$-frame predicted probability, the frame-level prediction at time $t$ is:
\begin{equation}
    \mathbf{\varphi}(f_{t}^{'}(x))=
    \left\{\begin{matrix}
1,&f_{t}^{'}(x)\cdot\mathbf{\phi}(f(x))\geq \beta \\ 
0,&otherwise
\end{matrix}\right.
\label{eq2}
\end{equation}

Without loss of generality, we take thresholds $\alpha=\beta=0.5$. According to Equation~\ref{eq2}, $\mathbf{\varphi}(f_{t}^{'}(x))$ depends on two sub-targets: $\mathbf{\phi}(f(x))$ (audio tagging) and $f_{t}^{'}(x)$ (boundary detection). As shown in Figure~\ref{fig3}, dense output loses recognition ability at a fine scale compared to the original scale. However, since dense output implies that the receptive field of the high-level feature map of the neural network is large enough to integrate contextual information, it tends to achieve better clip-level prediction. Therefore, when designing a neural network aiming at frame-level prediction, traditional methods choose a moderate temporal compression scale, which seeks a trade-off between audio tagging and boundary detection.

\begin{algorithm}[t]
\vskip 0.05in
\begin{small}
   \caption{Guided learning pseudocode.}
   \label{algorithm1}
\begin{algorithmic}
\REQUIRE $x_k = k^{th}$ training input sample
\REQUIRE $L = $ set of weakly-labeled training inputs
\REQUIRE $U = $ set of unlabeled training inputs 
\REQUIRE $y_k = $ label of $x_k$
\REQUIRE $S_\theta\left(x\right)=$ the neural network of the PS-model
\REQUIRE $T_{{\theta}^{'}}\left(x\right)=$ the neural network of the PT-model
\REQUIRE $J\left(t,z\right)=$ loss function 
\REQUIRE $\mathbf{\phi}(z)=$ prediction generation function
\ENSURE {$\theta, {\theta}^{'}$}
\FOR{$i=1 \rightarrow num\_epoches$}
\IF{$i>start\_epoch$}
\STATE $a\leftarrow 1-{\gamma}^{i-start\_epoch}$ \hfill$\triangleright$ the weight of $L^{'}_{unsupervised}$
\ELSE
\STATE $a\leftarrow 0$
\ENDIF
\FOR{each minibatch $\mathbf{\ss}$}
\STATE $L_{lPS}\leftarrow\frac{1}{\left |\mathbf{\ss}\right |} \left[
\sum_{x_k \in {L \cap \mathbf{\ss}}} J\left(y_k,S_{\theta}(x_k)\right)\right]$
\STATE $L_{lPT}\leftarrow\frac{1}{\left |\mathbf{\ss}\right |} \left[
\sum_{x_k \in {L \cap \mathbf{\ss}}} J\left(y_k,T_{\theta^{'}}(x_k)\right)\right]$

\STATE $L_{unsupervised}\leftarrow\frac{1}{\left |\mathbf{\ss}\right |} \left[
\sum_{x_k \in {U \cap \mathbf{\ss}}} J\left(\mathbf{\phi}((T_{\theta^{'}}(x_k)),S_{\theta}(x_k)\right)\right]$
\STATE $L^{'}_{unsupervised}\leftarrow\frac{1}{\left |\mathbf{\ss}\right |} \left[
\sum_{x_k \in {U \cap \mathbf{\ss}}} J\left(\mathbf{\phi}(S_{\theta}(x_k)),T_{\theta^{'}}(x_k)\right)\right]$
\STATE $loss=L_{lPS}+L_{lPT}+L_{unsupervised}+a\cdot L^{'}_{unsupervised}$
\STATE update $\theta, {\theta}^{'}$\hfill$\triangleright$ update network parameters
\ENDFOR
\ENDFOR
\end{algorithmic}
\end{small}
\end{algorithm}
Instead of seeking this trade-off, we design a promising student model (PS-model) with a relatively small temporal compression scale (even no compression), which is promising to achieve better boundary detection performance. Since we expect to design another model better at the direct training target to guide it to learn using unlabeled data, we design a professional teacher model (PT-model) with a larger time compression scale, which is more professional at audio tagging.
\abovedisplayskip 0.07in
\belowdisplayskip 0.07in

\subsection{The learning process}
\label{sec:Learning process}
In this section, we introduce the end-to-end learning process we proposed, in which the PT-model guides the PS-model to learn, as shown in Algorithm~\ref{algorithm1}.

Assuming that $S_{\theta}(x)$ and $T_{{\theta}^{'}}(x)$ are the clip-level predicted probabilities of the PS-model (with trainable parameters $\theta$) and the PT-model (with trainable parameters ${\theta}^{'}$) respectively, and $y$ is the groundtruth of input $x$. Then we force the PS-model to learn from unlabeled data with the tags that the PT-model predicts. Since the PT-model is explicitly superior to the PS-model, the audio tagging performance of the PS-model can be improved with the following loss:
\begin{equation}
    L_{unsupervised}=J(\mathbf{\phi}(T_{{\theta}^{'}}(x)),S_{\theta}(x))
\end{equation}
where $\mathbf{\phi}(T_{{\theta}^{'}}(x))$ denotes the clip-level prediction of the PT-model. Here, we take Cross Entropy as the loss function $J$.

To guarantee the basic ability of the PT-model, we utilize weakly-labeled data to update its weights with a supervised loss. To prevent the PS-model from learning too many noisy tags generated by the PT-model in the early stage and drifting away, we employ a supervised loss to update the PS-model. Then the combination of these two supervised losses is:
\begin{equation}
L_{supervised}=J(y,T_{{\theta}^{'}}(x))+J(y,S_{\theta}(x))
\end{equation}

We argue that as the training progresses, the capability of the PS-model on clip-level prediction gradually approaches the PT-model. Therefore, we can utilize unlabeled data with the tags from the PS-model to fine-tune the PT-model, which encourages the output probabilities of the PT-model to be insensitive to variations in the directions of the low-dimensional manifold just like the fine-tuning phase in Pseudo-Label \cite{lee2013pseudo}. Then the unsupervised loss is:
\begin{equation}
    L^{'}_{unsupervised}=J(\mathbf{\phi}(S_{\theta}(x)),T_{{\theta}^{'}}(x))
\end{equation}
where $\mathbf{\phi}(S_{\theta}(x))$ denotes the clip-level prediction of the PS-model. Therefore, the whole training process is as follows: 

At the beginning of the training, the PT-model simply updates with the supervised loss due to the poor performance of the PS-model. The fact that the superiority of the PT-model is guaranteed by the design of model architecture enables the PS-model to keep an unsupervised loss all the time. Therefore, in the first $s$ epochs, the loss we employ is:
\begin{equation}
L_{first}=L_{supervised}+L_{unsupervised}
\end{equation}

After $s$ epochs, the PT-model is considered to be a little stable and it begins to keep another loss to fine-tune itself. In this phase, the loss we employ is:
\begin{equation}
L_{second}=L_{first}+a\cdot L^{'}_{unsupervised}
\end{equation}

Here, the increasing factor $a$ is expected to be relatively small and increase slowly during training. If $a$ is too large, the PT-model will be effected by too-much noisy tags generated by the PS-model and drift away. Therefore, we take $a=1-{\gamma}^{e-s}$ in our experiments, where $e$ is the current epoch and $\gamma$ is a hyperparameter between 0.99 and 0.999.

\section{Experiments}
\label{sec:experiment}
\subsection{Data set}
The dataset we utilize is from the DCASE 2018 Task 4 \cite{Serizel2018}, which is divided into 5 subsets: weakly-labeled training set (1578 clips), unlabeled in domain training set (14412 clips), unlabeled out of domain training set (39999 clips), strongly-labeled validation set (288 clips) and strongly-labeled test set (880 clips). We take the combination of the weakly-labeled and unlabeled in domain training set as our training set.

\subsection{Model architecture}
\begin{figure}[t]
\vskip -0.05in
\begin{minipage}{0.44\linewidth}
  \centering
  \centerline{\includegraphics[width=\linewidth]{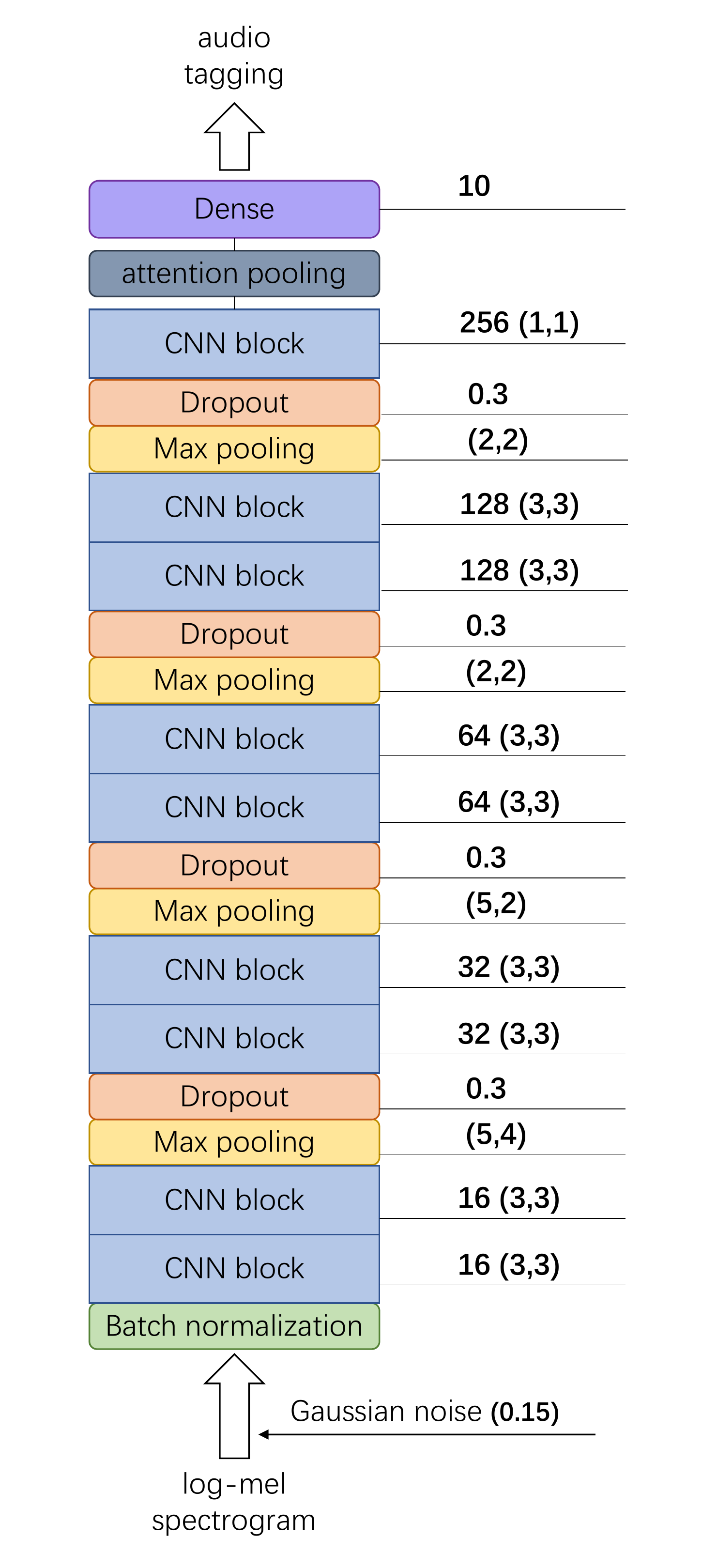}}
  \centerline{(a) PT-model}\medskip
\end{minipage}
\hfill
\begin{minipage}{0.44\linewidth}
\begin{minipage}{\linewidth}
  \centering
  \centerline{\includegraphics[width=\linewidth]{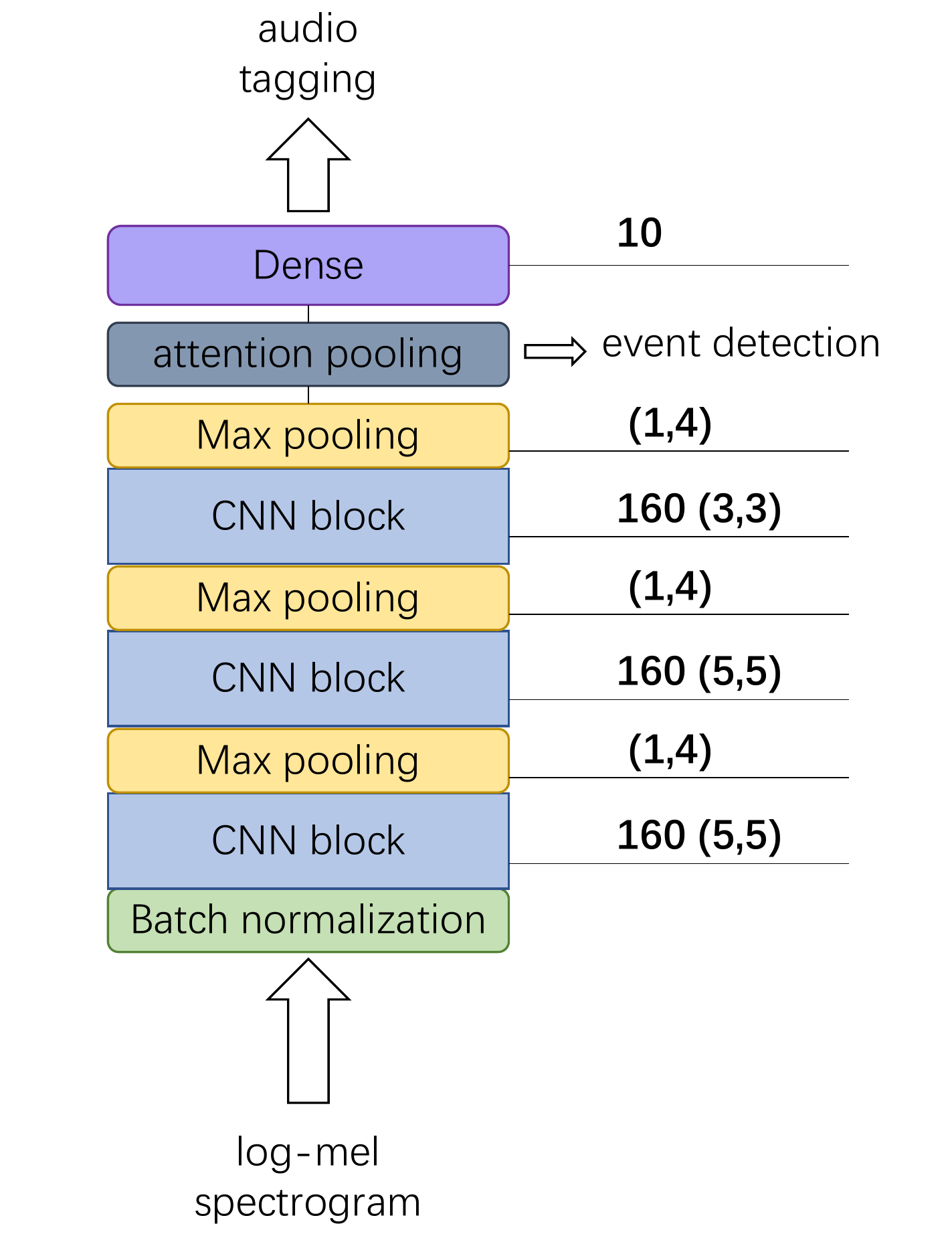}}
  \centerline{(b) PS-model}\medskip
\end{minipage}
\begin{minipage}{\linewidth}
  \centering
  \centerline{\includegraphics[width=\linewidth]{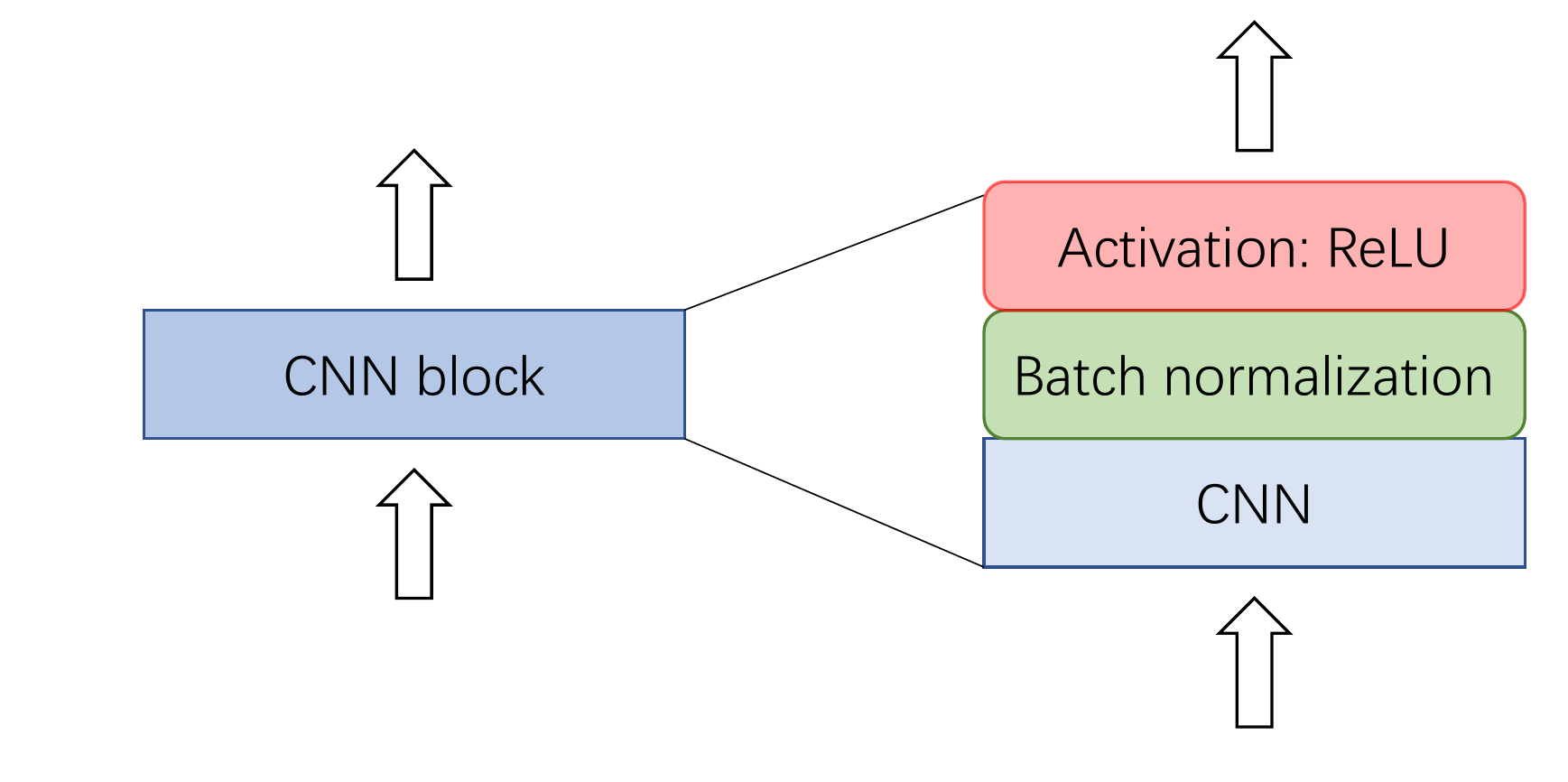}}
  \centerline{(c) CNN block}\medskip
\end{minipage}
\end{minipage}
\vskip -0.08in
\caption{The model architectures.}
\vskip -0.16in
\label{fig4}
\end{figure}

As shown in Figure~\ref{fig4}, the architecture of the PT-model comprises several components: a batch normalization input layer, 4 CNN modules, a CNN block, an attention pooling module and a dense layer. Each CNN block comprises a convolutional layer, a batch normalization layer and a ReLU activation layer. Each CNN module comprises a CNN block, a Max pooling layer and a dropout layer. The architecture of the PS-model comprises a batch normalization input layer, 3 CNN modules, an attention pooling module and a dense layer. The CNN modules in PS-model does not comprise any dropout layer. There is no temporal compression in the PS-model. The total number of trainable parameters of the PT-model is $332364$ and that of the PS-model is $877380$. The input augmentation is implemented by Gaussian noise ($\delta=0.15$) on the input layer of the PT-model. We employ 64 log mel-bank magnitudes which are extracted from 40 ms frames with $50\%$ overlap, thus each 10-second audio clip is transformed into 500 frames. During post-processing, a group of median filters with adaptive window size is utilized for smoothing the prediction.
\subsection{Training and evaluation}
The models are trained with a mini-batch of 64 clips for 100 epochs. All the experiments are repeated $30$ times. The average performances and the best performances within these $30$ experiments are reported. All the models are evaluated using the sed\_eval package to compute event-based measure with a 200ms collar on onsets and a 200ms / $20\%$ of the events length collar on offsets. We report clip-level macro $F_1$ score \cite{mesaros2016metrics} to measure the performance of models on audio tagging. We release our code at https://github.com/Kikyo-16/Sound\_event\_detection.

\begin{table}[t]
  \caption{The average performances of models}
  \vskip 0.1in
  \label{table1}
  \centering
\begin{tabular}{lcc}
\toprule
\textbf{Model}
&\textbf{\textbf{Audio tagging}}
&
\textbf{Event-based $F_1$}\\
\midrule
The $\mathbf{1^{st}}$ place&-&-\\
Mean Teacher&$0.609\pm0.003$&$0.300\pm0.004$\\
\hline
\multicolumn{2}{l}{Weakly-labeled-only}\\
PS-model&$0.582\pm0.004$&$0.247\pm0.004$\\
PT-model&$0.611\pm0.003$&$0.112\pm0.002$\\
\hline
GL-1&$0.637\pm0.004$&$0.340\pm0.005$\\
GL-0.999&$\mathbf{0.652\pm0.004}$&$0.360\pm0.007$\\
GL-0.998&$0.648\pm0.005$&$0.360\pm0.008$\\
GL-0.997&$0.641\pm0.003$&$\mathbf{0.363\pm0.005}$\\
GL-0.999*&$0.611\pm0.003$&$0.335\pm0.005$\\
\bottomrule 
\end{tabular}
\vskip -0.16in
\end{table}

\subsection{Results}
We set start epoch $s=5$. As shown in Table~\ref{table1}, GL-0.997 (Guided Learning with $\gamma=0.997$) achieves the best average performance with an event-based $F_1$ score of $0.363$. As shown in Table~\ref{table2}, GL-0.999 achieves an event-based $F_1$ score of $39.5\%$ (the best result within these $30$ experiments), outperforming the first place \cite{jiakai2018mean} in the challenge by $7.1$ percentage points. The first place in the challenge employs Mean Teacher \cite{tarvainen2017mean} with a well-designed CRNN model and achieves an event-based $F_1$ score of $32.4\%$, which is to our knowledge the SOTA performance on the dataset so far.

\textbf{Extra baseline} We also combine Mean Teacher \cite{tarvainen2017mean} with our weakly-labeled SED system to be another baseline. Since Mean teacher requires two models with the same model architecture, we employ the architecture of the PS-model described in Figure~\ref{fig4} as their architecture. As shown in Table~\ref{table1} and~\ref{table2}, GL outperforms Mean Teacher both on audio tagging and event-based $F_1$ score on the dataset. 

\textbf{Weakly-labeled learning} The weakly-labeled-only PT-model achieves an average macro $F_1$ score of $0.611$ on audio tagging and outperforms the PS-model, which exactly meets our expectation. This is because the temporal size of the feature map of the PT-model is compressed by several pooling layers step by step. Finally, the receptive field of the high-level feature map is large which makes it easy to integrate the contextual information. On the contrary, the receptive field of every-frame high-level feature representation of the PS-model is only $2.2\%$ $(11/500)$ of its original temporal scale. The gap between the capability of the two models to integrate the contextual information explains why the PT-model achieves better audio tagging performance. On the other hand, the PS-model achieves an average event-based $F_1$ score of $0.238$ on SED, outperforming the PT-model by $13.4$ percentage. Obviously, the PT-model lost the ability to see finer information due to its temporal scale compression. 

\begin{table}[t]
  \caption{The performances with the best event-based $F_1$ score}
  \vskip 0.1in
  \label{table2}
  \centering
\begin{tabular}{lcc}
\toprule
\textbf{Model}
&\textbf{\textbf{Audio tagging}}
&
\textbf{Event-based $F_1$}\\
\midrule
The $\mathbf{1^{st}}$ place&-&0.324\\
Mean Teacher&$0.625$&$0.324$\\
\hline
\multicolumn{2}{l}{Weakly-labeled-only}\\
PS-model&$0.600$&$0.267$\\
PT-model&$0.627$&$0.124$\\
\hline
GL-1&$0.653$&$0.371$\\
GL-0.999&$\mathbf{0.673}$&$\mathbf{0.395}$\\
GL-0.998&$0.671$&$0.392$\\
GL-0.997&$0.657$&$0.388$\\
GL-0.999*&$0.632$&$0.367$\\
\bottomrule 
\end{tabular}
\vskip -0.16in
\end{table}
\textbf{The effect of GL} All the GL models outperform weakly-labeled-only PT-model and PS-model both on audio tagging and event-based $F_1$ score. When $\gamma=1$, the weight of $L^{'}_{unsupervised}$ is always 0 and the PT-model does not learn from the unlabeled data. Then the fact that GL-1 outperforms weakly-labeled-only PT-model on audio tagging indicates that the unlabeled data does not only help the PS-model to catch up with the PT-model on audio tagging but also helps it to exceed the original PT model without GL.  When $\gamma$ decreases to $0.999$, the PS-model in GL-0.999 achieves the best performance on audio tagging, which denotes that the PT-model in GL-0.999 is fine-turned by $L^{'}_{unsupervised}$. However, when $\gamma$ decreases to $0.998$, the weight of $L^{'}_{unsupervised}$ increases more rapidly and the audio tagging performance of the PS-model in GL-0.998 starts to decline. 
To highlight the effect of the PT-model in GL, we make the PT-model share the same architecture of the PS-model in GL-0.999 and show the results as GL-0.999* in the tables. Comparing the performance of GL-0.999 and GL-0.999* in Table~\ref{table1}, GL-0.999 performs better. Since there is no performance gap on audio tagging between the two models in GL-0.999*, the result shows that the better ability on audio tagging of the PT-model does contribute to the better performance of GL. In addition, the trainable parameters of the PT-model are always much fewer than that of the PS-model, which greatly improves the training efficiency.

\section{Conclusions}
\label{sec:conclusions}
We investigate the two sub-targets implied in weakly-labeled SED and relate them to the design of model architecture, based on which we proposed a semi-supervised learning method named Guided Learning to optimize weakly-labeled SED system using unlabeled data. In addition, some other existing researches simply focus on audio tagging and the proposed approach provides a direct way to reuse those SOTA model to a weakly-labeled SED system.
\section{ACKNOWLEDGMENT}
This work is partly supported by Beijing Natural Science Foundation (4172058).

\vfill\pagebreak

\bibliographystyle{IEEEbib}
\bibliography{strings}

\end{document}